\documentclass[runningheads]{llncs}

\usepackage{eccv}

\usepackage{eccvabbrv}

\usepackage{graphicx}
\usepackage[table]{xcolor}
\usepackage{multirow}
\usepackage{booktabs}
\usepackage{xcolor}
\usepackage{pifont} 
\usepackage{wrapfig}
\usepackage{amssymb}   
\usepackage[accsupp]{axessibility}  

\usepackage{hyperref}

\usepackage{orcidlink}

\begin{document}

\title{TruthLens: Object Hallucination Detection via Self-Evaluating Truthfulness Scores in LVLMs} 

\titlerunning{TruthLens}

\author{Yanqi Wu\inst{1,2,5}\orcidlink{0009-0000-3544-2325} \and
Runhe Lai\inst{1,2,5}\orcidlink{0009-0001-7843-6512} \and
Xinhua Lu\inst{1,2,5}\orcidlink{0009-0001-6222-4005} \and
Qichao Chen\inst{3}\orcidlink{0009-0006-2140-9042} \and
Zhiping Zhou\inst{1,2,5}\orcidlink{0009-0001-7152-6662} \and
Jia-Xin Zhuang\inst{4}\orcidlink{0000-0001-9287-4263} \and
Weijiang Yu$^\star$\inst{1}\orcidlink{0000-0002-7449-3093} \and
Ruixuan Wang\thanks{Corresponding author.}\inst{1,2,5}\orcidlink{0000-0002-8714-0369}
}

\authorrunning{Wu et al.}

\institute{
School of Computer Science and Engineering, Sun Yat-sen University, Guangzhou, China \and
Peng Cheng Laboratory, Shenzhen, China \and
University of Nottingham Malaysia, Semenyih, Malaysia \and
Hong Kong University of Science and Technology, Hong Kong, China \and
Key Laboratory of Machine Intelligence and Advanced Computing, MOE, Guangzhou, China \\
\email{\{wuyq268, lairh5, luxh55\}@mail2.sysu.edu.cn},
\email{hcxqc1@nottingham.my}, \email{jzhuangad@cse.ust.hk}, \email{\{yuwj39,wangruix5\}@mail.sysu.edu.cn},
}

\maketitle

\begin{abstract}
    Despite the remarkable progress of large vision language models (LVLMs), object hallucination remains a fundamental challenge that hinders their trustworthy deployment. 
    A key finding motivates our work: real and hallucinated object tokens are clearly separable in hidden representations, yet this separability is largely lost at the language-modeling (LM) head. We propose TruthLens, a self-evaluation framework that teaches the LM head to expose a per-object truthfulness signal without any auxiliary model or additional inference cost. Concretely, a rarely-used special token is repurposed as a reference token. For each object-token position, we extract the log-probability assigned to this special token by the LM head, and define its difference from a predefined constant as the truthfulness score.
    The model is then fine-tuned with an MSE objective that drives scores toward 1 for real objects and 0 for hallucinated ones, while a divergence constraint preserves the original generation capability.
    Despite being trained on only a limited set of object categories, TruthLens generalizes effectively to benchmarks with substantially larger label spaces. Extensive experiments across multiple LVLMs demonstrate state-of-the-art performance; notably, on Qwen2.5-VL-7B, TruthLens outperforms the previous best method on MS-COCO by over 17\% in AUROC. Our code is available at \url{https://github.com/wyqstan/TruthLens}.

    \keywords{Object hallucination detection \and Large vision language model}
\end{abstract}

\section{Introduction}
\label{sec:intro}
Large vision-language models (LVLMs)~\cite{llava,qwen25vl,oneVision15,mPLUG-Owl3} have achieved remarkable success in image captioning, visual question answering, and complex multimodal reasoning by integrating visual perception with the language understanding capabilities of large language models (LLMs). Despite their impressive performance, LVLMs frequently suffer from \textit{object hallucination} (OH)~\cite{chair,pope}-- 
generating mentions of objects absent from the input image
--which poses a serious challenge to their reliability in real-world deployment.

\begin{figure}[t]
    \centering
    \includegraphics[width=\linewidth]{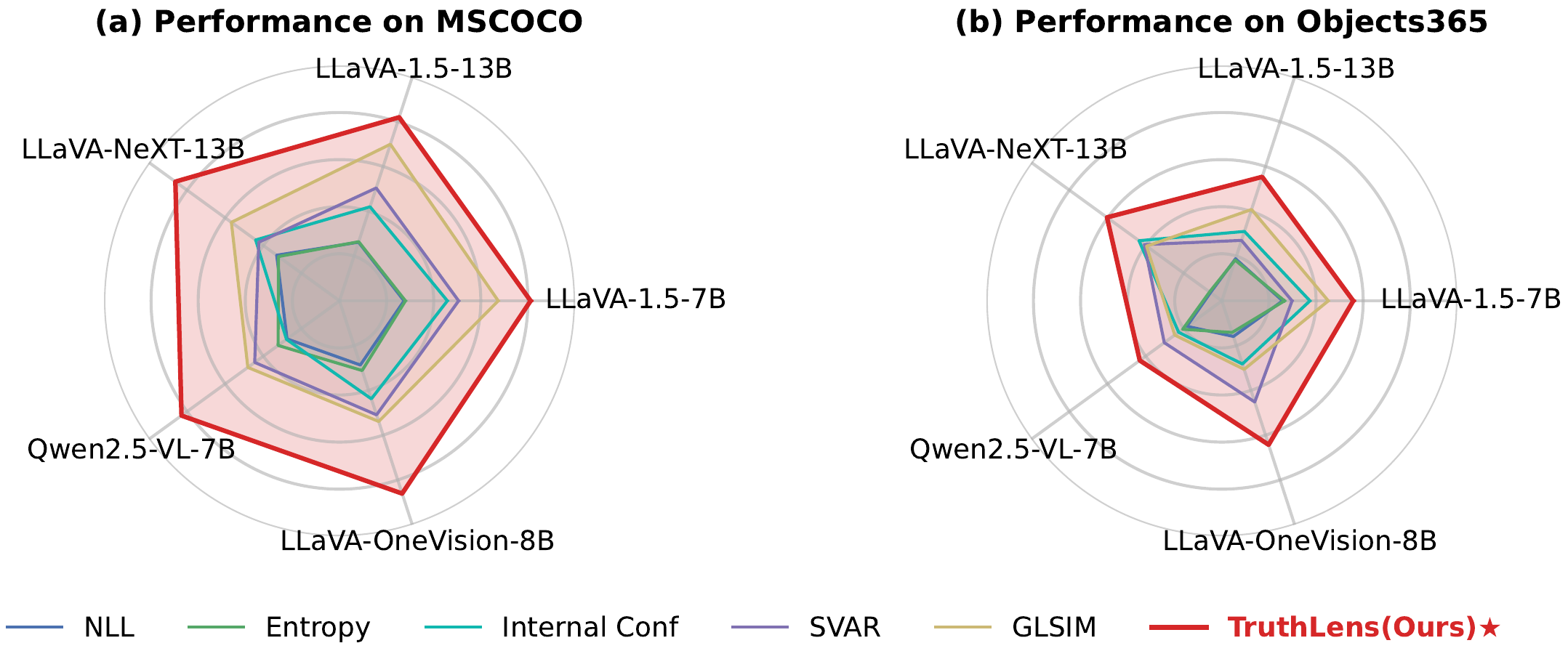}
    \caption{Hallucination detection performance of five open-source LVLMs~\cite{llava15,llavanext,qwen25vl,oneVision15} using six different scoring metrics on MSCOCO~\cite{MSCOCO} and Object365~\cite{Object365} datasets.
 }
\label{fig:performance_overview}
\end{figure}

Existing LVLM OH detection methods~\cite{SVAR,IC,Contextual_Lens} predominantly focus on visual grounding verification by explicitly analyzing image tokens or vision-language interactions. While effective, these approaches largely overlook the intrinsic properties of LLMs, where hallucination is often studied as a language-side phenomenon related to internal confidence and uncertainty~\cite{WhyLanguageModelsHallucinate,uncertaintyquantificationconfidencecalibration,trustmeimwrong}. 
Recent studies on LLMs~\cite{kadavath2022languagemodelsmostlyknow,li2023inference,chen2024insidellmsinternalstates,orgad2025llmsknowshowintrinsic} have shown that hallucination-related signals are implicitly encoded in internal representations: hidden states often carry information about the model's uncertainty or the factual correctness of generated tokens. Since modern LVLMs generate responses autoregressively in the language space---with visual information injected through cross-modal conditioning---they can be viewed as LLMs whose token representations are additionally conditioned on visual inputs. This perspective raises a natural question:
\begin{center}
    \textit{Do LVLMs also encode truthfulness signals in their language-side representations, and can these signals be surfaced for hallucination detection?}
\end{center}

To investigate this question, diagnostic analyses are conducted using two complementary probes: token-level negative log-likelihoods reflecting the LM head’s prediction confidence, and linear discriminant analysis (LDA)~\cite{LDA} applied to hidden representations that measures intrinsic feature separability. We use these signals to compare the distinguishability of real and hallucinated tokens at the final output layer and the intermediate feature layer. 
As shown in Fig.~\ref{fig:LDA_NLL}, real and hallucinated object tokens are clearly separable in the hidden feature space,
yet this separability is substantially attenuated in the output probability distributions produced by the LM head. This gap between hidden-state discriminability and output-level ambiguity points to a concrete opportunity: if the LM head can be taught to preserve and expose these discriminative signals, the model itself can serve as its own hallucination detector.

Drawing inspiration from LASER~\cite{LASER}, which quantifies the reasoning reward of a complete solution via the log-probability of a designated special token, we propose \textbf{TruthLens}---a self-evaluation framework that extends this idea to the \textit{token level}.
A rarely-used special token is repurposed as a probing signal. At each prediction step corresponding to an object token, we measure the log-probability assigned to this special token and subtract a predefined constant to obtain a truthfulness score for that object token.
During training, multiple captions are sampled for each image, and object tokens in these captions are labeled as real or hallucinated. A mean squared error (MSE) loss then drives the truthfulness scores of real tokens toward~1 and those of hallucinated tokens toward~0, while a divergence constraint prevents degradation of the model's original generation capability. At inference time, the truthfulness score is used directly for OH detection, introducing zero inference overhead.


As shown in Fig.~\ref{fig:performance_overview}, we evaluate our method across multiple LVLMs and observe consistent improvements over existing OH detection approaches. Although our training annotations cover only a limited set of object categories, the proposed method generalizes effectively to benchmarks with substantially larger label spaces, demonstrating strong generalization capability.
Our main contributions can be summarized as follows:

\begin{itemize}
\item We uncover a key phenomenon: real and hallucinated object tokens are clearly separable in the hidden feature space, yet this separability is largely attenuated in the output distributions produced by the LM head.
\item We define the truthfulness score of an object token as the log-probability assigned to a designated special token by the model’s output distribution at its prediction step, minus a predefined constant. This allows the LM head to explicitly expose a confidence signal of factual correctness.
\item We conduct extensive experiments across multiple LVLMs and benchmarks, demonstrating the effectiveness, robustness, and strong generalization capability of the proposed method for OH detection.
\end{itemize}

\section{Related Work}

\noindent \textbf{Large Vision-Language Models.}
The integration of advanced open-source LLMs, such as LLaMA~\cite{llama,llama2}, Vicuna~\cite{vicuna}, and Qwen~\cite{qwen25,qwen2}, has significantly advanced large vision--language models (LVLMs) in complex vision--language tasks. Most LVLMs follow a three-component architecture, consisting of a vision encoder, a modality connector, and a pretrained language model. Specifically, the vision encoder, often based on CLIP~\cite{clip}, extracts visual features, while the connector projects them into the textual embedding space for joint reasoning with textual prompts. For example, LLaVA~\cite{llava,llava15} adopts an MLP projector, whereas recent models, such as mPLUG-Owl3~\cite{mPLUG-Owl3}, Qwen2.5-VL~\cite{qwen25vl}, and LLaVA-OneVision-1.5~\cite{oneVision15}, introduce more advanced visual fusion mechanisms, including Q-Former~\cite{blip2} and DeepStack~\cite{deepstack}. Despite their strong performance, LVLMs remain prone to severe object hallucinations. In this work, we systematically analyze and address this issue through targeted hallucination detection.

\noindent \textbf{Object Hallucination Detection.}
Object hallucination (OH) refers to the phenomenon in which LVLMs generate text that mentions objects absent from the input image. This issue poses significant risks to the reliability of LVLM-based systems, particularly in safety-critical scenarios such as medical imaging and autonomous driving.
Existing OH detection approaches can be broadly categorized into two groups: methods that rely on auxiliary models and those that exploit internal signals of LVLMs. In the first category, GAIVE~\cite{GAIVE} employs a stronger LVLM as a teacher model to evaluate student outputs, while HaLEM~\cite{HaLEM} fine-tunes an LLM to score LVLM generations. Although effective, these approaches introduce additional computational overhead by requiring extra models at inference time, thereby increasing latency.
The second line of work detects hallucinations directly from the internal representations of LVLMs. LURE~\cite{LURE} leverages the negative log-likelihood (NLL) of object token probabilities for detection. Internal Confidence (IC)~\cite{IC} assesses object existence by taking the maximum object-token probability across image embeddings. Summed Visual Attention Ratio (SVAR)~\cite{SVAR} quantifies the contribution of visual information by measuring the proportion of attention a generated token allocates to image embeddings. GLSIM~\cite{GLSIM} integrates patch-level visual cues with global representations of both the image and the prompt to construct a holistic hallucination detector.
InsLen\cite{inslen} introduces a plug-and-play, training-free OH detection method for MLLMs by leveraging instruction token embeddings to capture implicit visual cues and calibrating local visual evidence with contextual consistency.
In contrast, we approach OH detection from a different perspective. We observe that the hidden states of real and hallucinated object tokens exhibit clear separability in the feature space
, while this separability is largely attenuated in the probability distributions produced by the LM head. Motivated by this observation, we fine-tune the model with only a small amount of annotated data, enabling the LM head to explicitly expose this intrinsic separability for more reliable hallucination detection.
\section{Method}

\subsection{Preliminaries}
\textbf{Object Hallucination Detection.}
Consider a LVLM that takes an image $I$ together with an instruction $X$ as input and generates a token sequence $Y={y_1, \dots, y_T}$. Among the generated tokens, we focus on those $y_t$ corresponding to object mentions. The goal of OH detection is to determine, for each such object token $o=y_t$, whether it is visually grounded in the input image $I$ or instead represents a hallucinated concept.
To this end, we design a scoring function $S(o)$ for each object token to quantify the model’s confidence regarding its factual correctness. Based on this score, we assign a binary decision as follows:

\begin{equation}
\label{Eq:decision_rule}
\small
D(o) =
\begin{cases}
\text{Hallucination}, & S(o) \le \mu, \\
\text{Truth}, & S(o) > \mu,
\end{cases}
\end{equation}
where $D(o)$ denotes the binary detection decision for an object token $o$, and $\mu$ is a fixed threshold that maps the confidence score to a discrete detection outcome.

\begin{figure}[t]
    \centering
    \includegraphics[width=\linewidth]{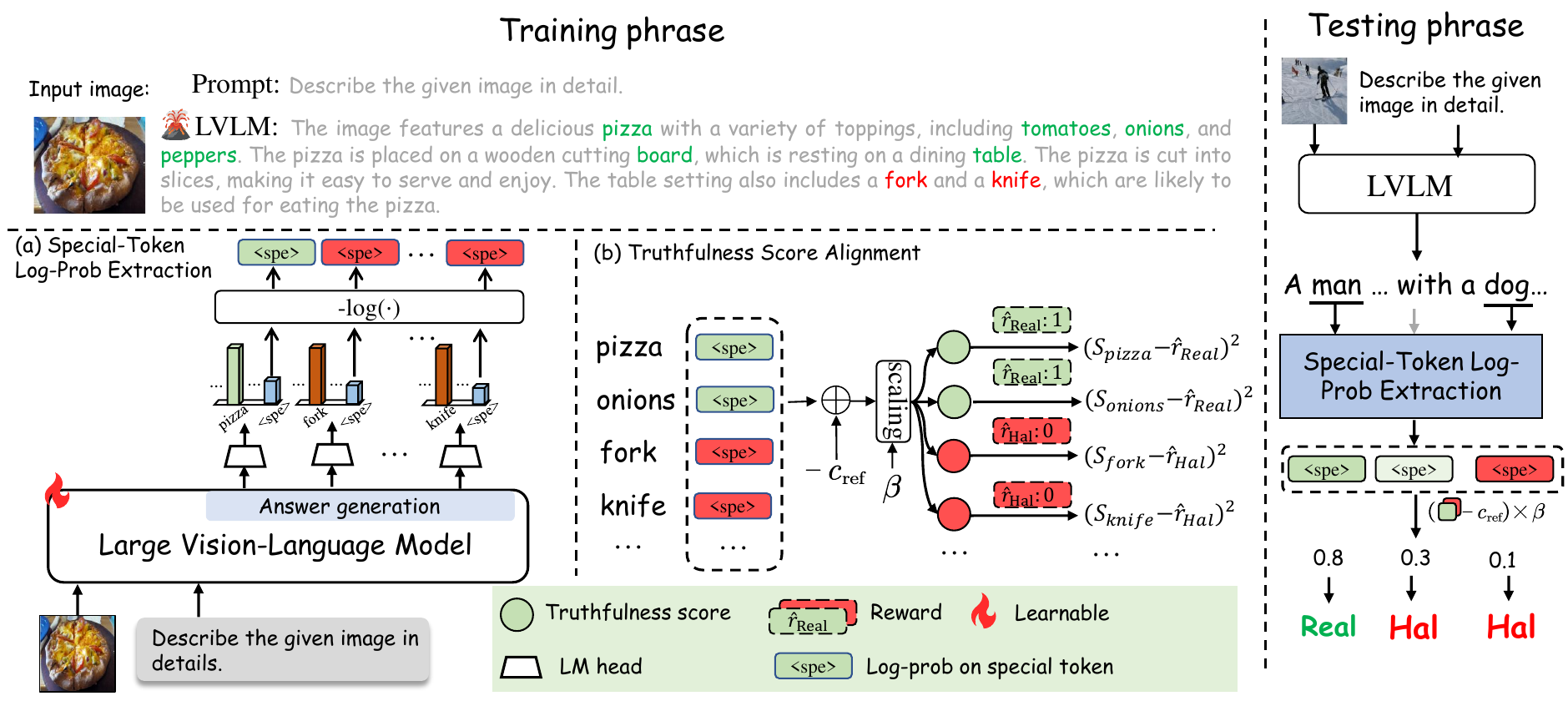}
    \caption{Overview of the proposed self-evaluation framework. The truthfulness score is formulated as the difference between the log-probability of a designated special token, predicted at the corresponding token position, and a constant. During training, an MSE loss aligns this score with binary rewards (1 for real, 0 for hallucinated). At inference, the score is directly leveraged for hallucination detection without auxiliary models.
 }

\label{fig:framework}
\end{figure}

\subsection{Overview}

As illustrated in Fig.~\ref{fig:framework}, we introduce a token-level self-evaluation truthfulness score that enables LVLMs to assess whether their generated tokens correspond to hallucinated objects. Our approach is motivated by a key observation: although the hidden states of real and hallucinated object tokens are clearly separable in the feature space, this separability cannot be effectively exposed by the LM head of LVLMs (Section~\ref{motivation}).
To address this issue, we define the truthfulness score of an object token as the log-probability assigned to a designated special token by the model’s output distribution at its prediction step, minus a predefined constant. During training, object tokens associated with real and hallucinated objects are supervised using a simple MSE loss, encouraging their truthfulness scores to approach 1 and 0, respectively.
At inference time, OH detection is performed directly based on these token-level truthfulness scores (Section~\ref{self_evaluation_score}).

\subsection{Motivation}
\label{motivation}
Recent studies~\cite{orgad2025llmsknowshowintrinsic,bui2025correctness} have demonstrated that, in LLM-generated responses, the hidden representations at token positions directly corresponding to the answer encode substantial information about answer correctness.
By analogy, in OH detection, the token positions corresponding to generated object mentions can be viewed as critical decision points that determine the factual consistency of the output. We therefore hypothesize that the hidden representations at these object-token positions similarly encode informative signals regarding whether the mentioned objects are grounded in the visual input or hallucinated.

\begin{figure*}[t]
    \centering
    
    \begin{subfigure}[t]{0.48\textwidth}
        \centering
        \includegraphics[width=\linewidth]{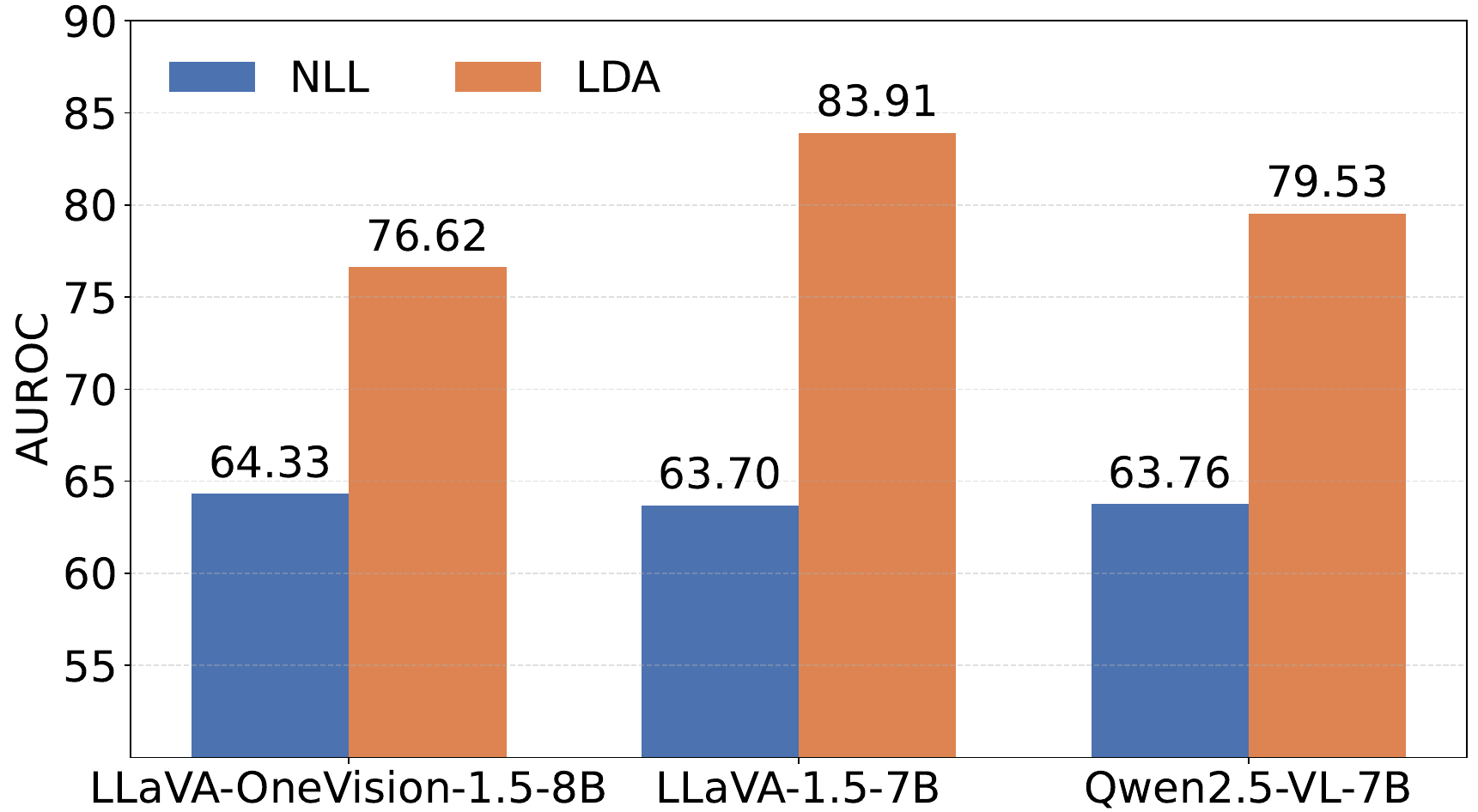}
        \caption{LDA vs. NLL Comparison}
        \label{fig:LDA_NLL_comparison}
    \end{subfigure}
    \hfill
    \begin{subfigure}[t]{0.48\textwidth}
        \centering
        \includegraphics[width=\linewidth]{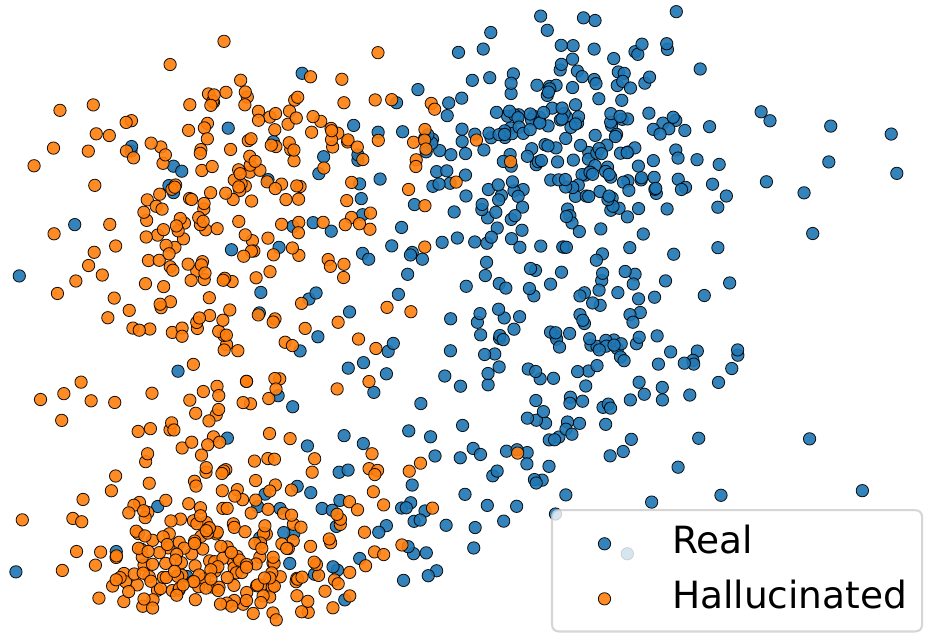}
        \caption{LDA on LLaVA-1.5-7B}
        \label{fig:llava_LDA}
    \end{subfigure}

    \begin{subfigure}[t]{0.48\textwidth}
        \centering
        \includegraphics[width=\linewidth]{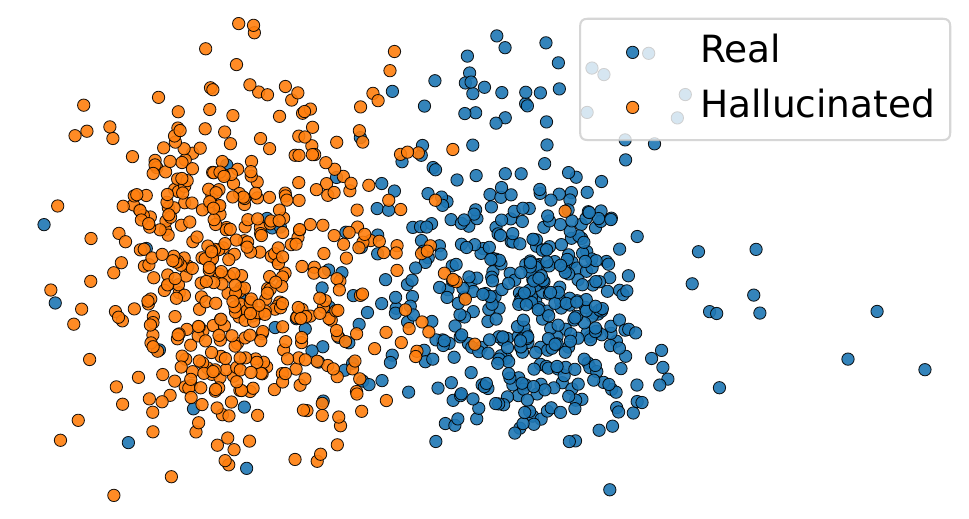}
        \caption{LDA on Qwen2.5-VL-7B}
        \label{fig:qwen_LDA}
    \end{subfigure}
    \hfill
    \begin{subfigure}[t]{0.48\textwidth}
        \centering
        \includegraphics[width=\linewidth]{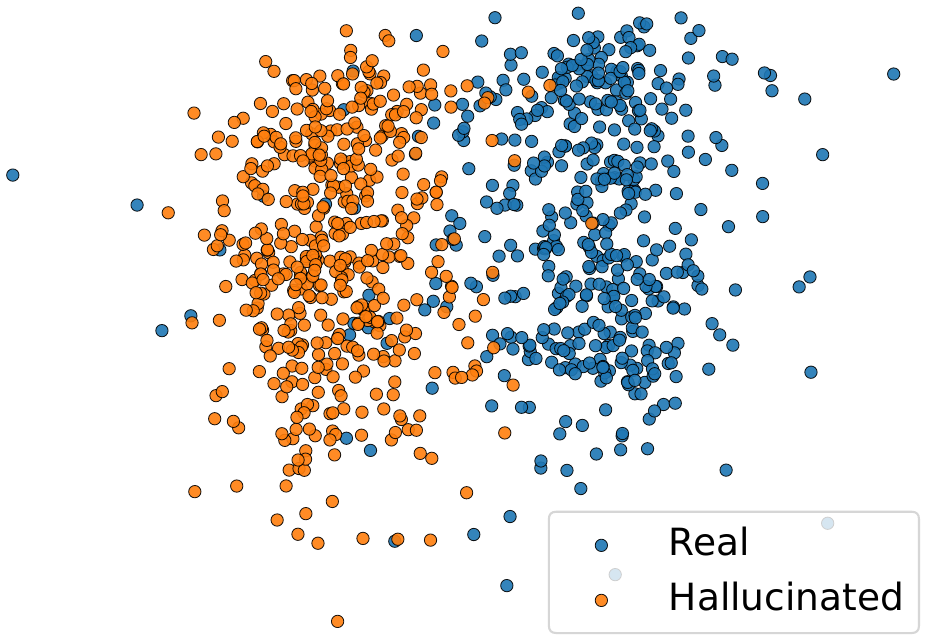}
        \caption{LDA on LLaVA-OneVision-1.5-8B}
        \label{fig:oneVision_LDA}
    \end{subfigure}

    \caption{Comparison of hallucination detection performance and LDA-based visualization on MSCOCO across different LVLM architectures.}
    \label{fig:LDA_NLL}
\end{figure*}

To validate this hypothesis, we conduct experiments on three representative LVLMs: LLaVA-1.5-7B~\cite{llava15}, Qwen2.5-VL-7B~\cite{qwen25vl}, and LLaVA-OneVision-1.5-8B~\cite{oneVision15}.
Specifically, we randomly sample 5,000 images from the MSCOCO dataset and prompt each model to generate captions. From the generated outputs,
we extract the hidden states from the last Transformer layer
corresponding to real object tokens and hallucinated object tokens, respectively.
We then perform LDA~\cite{LDA} on these features, using 80\% of the samples for training and the remaining 20\% for evaluation. As shown in Fig.~\ref{fig:LDA_NLL}, the hidden representations of real and hallucinated object tokens exhibit clear separability. In contrast, the model’s predicted probabilities for these tokens—measured by their negative log-likelihoods (NLLs)—do not demonstrate comparable discriminative power (Fig.~\ref{fig:LDA_NLL_comparison}). This discrepancy indicates that the discriminative truthfulness signals encoded in the model’s intermediate hidden states are not fully propagated or preserved in the final output probabilities, meaning the model itself fails to effectively leverage these intrinsic cues for self-evaluation.
Motivated by this observation, we introduce a token-level truthfulness score for each object token and fine-tune the model in a way that preserves its original task performance, enabling the model to explicitly read out the truthfulness signals encoded in its internal representations.

\subsection{Self-Evaluation Truthfulness Score}
\label{self_evaluation_score}

During LVLM generation, tokens assigned high probabilities by the LM head primarily serve the core objective of language modeling. Directly modifying the predicted probabilities of these tokens for OH detection would therefore risk disrupting the model’s original generative capabilities.
In contrast, certain tokens consistently receive extremely low predicted probabilities and are rarely generated during inference, such as special tokens (e.g., image placeholders or unknown tokens). Slightly adjusting the probabilities of these low-probability tokens has minimal influence on the overall probability distribution and does not materially affect the generated outputs.
Therefore, we select a designated low-probability special token and slightly adjust its predicted probability so that it encodes the distinction between real and hallucinated object tokens.

Recently, LASER~\cite{LASER} introduces a formulation in which the true reasoning reward of a complete solution is quantified by the log-probability of a designated special token at the next-token prediction following the last response token.
Although this reward formulation is originally designed for complete model responses, we find that it remains effective when applied at the level of individual tokens. In OH detection, we define the reward $\hat{r}$ for an object token $o$ as
\begin{equation}
\label{reward}
\hat{r}(o, I) =
\begin{cases}
1, & \text{if object } o \text{ is present in image } I, \\
0, & \text{otherwise}.
\end{cases}
\end{equation}
The truthfulness score $S(o,X,I)$ of an object token is formulated as
\begin{equation}
\small
\label{truthfulness_score}
S(o,X,I) = \log \pi_\theta \!\left( z_c \mid X, I, y_{<o} \right) - c_{\mathrm{ref}},
\end{equation}
where $\pi_\theta$ denotes the fine-tuned LVLM in our framework, $z_c$ is the pre-defined special token, $X$ represents the textual prompt, $I$ is the input image, $y_{<o}$ denotes the sequence of tokens generated before token $o$, and $c_{\mathrm{ref}}$ is a predefined constant.

Our objective is to align the truthfulness score $S(o, X, I)$ of each object token with its target reward, where real object tokens are assigned 1 and hallucinated object tokens 0.
To achieve this goal, we fine-tune the model using a small set of images with dense object annotations. Specifically, for each image, we sample $n$ captions from the model and extract the object tokens from the generated outputs. When an object mention is tokenized into multiple sub-tokens, we retain only the first token for supervision. Analyses of alternative supervision positions are provided in Section~\ref{further_analysis}.
Based on the ground-truth annotations, these object tokens are categorized as either real or hallucinated. We then apply a MSE loss to align the predicted truthfulness score of each token with its corresponding target reward. The overall training objective is defined as follows:
\begin{equation}  
\small
\label{loss} 
L = \mathbb{E}_{x, I \sim D,\; y \sim \pi_\theta(\cdot \mid x,I),\; o \sim y} \Bigl[ \beta S(o,x,I) - \hat{r}(o, I) \Bigr]^2 + D_{\mathrm{KL}}\!\left( \pi_\theta(y \mid x,I)\,\|\, \pi_{ref}(y \mid x,I) \right). 
\end{equation}
where $\beta$ is a scaling hyperparameter used to control the gradient magnitude for stable training. The KL divergence constrains the deviation between the model $\pi_\theta$ and the reference model $\pi_{ref}$, which is a frozen copy of the initial model before fine-tuning and shares the same architecture as the model $\pi_\theta$, thereby preventing performance degradation caused by excessive distributional divergence.

During training, the numbers of real and hallucinated object tokens are inherently imbalanced, and their proportions may vary across batches. To prevent the truthfulness score from being biased toward the majority class, we adopt a class-level loss re-weighting strategy at each optimization step. At each training step, we count the numbers of real and hallucinated object tokens within the current batch, denoted as $N_r$ and $N_h$, and apply loss re-weighting as
\begin{equation}
\small
\label{reweight_loss}
\begin{aligned}
L
&= \frac{1}{N_r + N_h}
\sum_{x,I}
\sum_{y \sim \pi_\theta(\cdot \mid x,I)}
\sum_{o \in \mathcal{O}(y)}
\Bigl[
w_r \mathbf{1}_{\{\hat r(o,I)=1\}}
+ w_h \mathbf{1}_{\{\hat r(o,I)=0\}}
\Bigr] \\
&\quad \times
\Bigl(
\beta S(o,x,I)
- \hat r(o,I)
\Bigr)^2
+ D_{\mathrm{KL}}\!\left( 
\pi_\theta(y \mid x,I)
\,\|\, 
\pi_{ref}(y \mid x,I)
\right),
\end{aligned}
\end{equation}
where $w_r=\frac{N_r+N_h}{2N_r}$ and $w_h=\frac{N_r+N_h}{2N_h}$ are the re-weighting factors. and $\mathcal{O}(y)$ denotes the set of object tokens in the generated sequence $y$. This re-weighting strategy enables a more balanced self-evaluation of token-level truthfulness across real and hallucinated object tokens.

During inference, we compute the truthfulness score for each generated object token according to Eq.~\ref{truthfulness_score}. Each token is then classified as real or hallucinated by 
applying the decision rule in Eq.~\ref{Eq:decision_rule},
enabling OH detection without introducing additional models or incurring extra inference-time overhead.

\section{Experiments}

\subsection{Experimental setups}

\textbf{Datasets.} 
For OH detection, MS-COCO~\cite{MSCOCO} (80 categories) and Object365~\cite{Object365} (365 categories) serve as the evaluation benchmarks, following prior work~\cite{GLSIM}. During evaluation, a fixed prompt---\textit{``Describe the given image in detail.''}---is used as the textual input to the model. 
To further assess the generality of TruthLens, we further investigate object attribute hallucinations (OAH) in LVLMs by assessing the correctness of generated attribute descriptions. For this purpose, we adopt CLEVR~\cite{CLEVR} and SpatialMQA~\cite{SpatialMQA} as benchmark datasets for OAH detection. CLEVR includes questions that require reasoning over object attributes and relationships, such as color, material, and relative spatial positions. In contrast, SpatialMQA presents a more challenging scenario, emphasizing fine-grained reasoning about relative object locations.


\noindent \textbf{Models.} 
Five representative LVLMs are evaluated: LLaVA-1.5 (7B and 13B)~\cite{llava15}, LLaVA-NeXT-13B~\cite{llavanext}, Qwen2.5-VL-7B-Instruct~\cite{qwen25vl}, and LLaVA-OneVision-1.5-8B-Instruct~\cite{oneVision15}, spanning different architectures and model scales.

\noindent\textbf{Implementation Details.} In our experiments, for OH detection, we randomly sample 4,000 images from the MSCOCO training split for fine-tuning. 
For OAH detection, we randomly select 6,500 samples from the training split of CLEVR and use all training data from SpatialMQA, respectively.
We employ LoRA~\cite{lora} for parameter-efficient fine-tuning of the LVLMs. The LoRA rank $r$ is set to 8, and the scaling factor $\alpha$ for LoRA is set to 16. The learning rate is set to $5 \times 10^{-5}$ with a batch size of 16. For each training sample, we generate 8 responses via stochastic sampling to provide supervision signals. Following CHAIR~\cite{chair}, we extract object tokens from the generated descriptions and match their normalized forms against the ground-truth object classes of each image and their corresponding synonyms.
Regarding the designated special token, we use the <unk> token for the LLaVA series models. For Qwen2.5-VL-7B and LLaVA-OneVision-1.5-8B, we instead adopt the <|image\_pad|> token. For other hyperparameters, we set $\beta = 0.1$ and $c_{\text{ref}} = -28.0$. The negative value of $c_{\text{ref}}$ is due to the log-probability scale, where token probabilities lie in $(0,1)$ and thus their logarithms are negative. More detailed analysis of these hyperparameters is provided in Section~\ref{further_analysis}.
During inference, we randomly sample 5,000 test instances from MSCOCO, Object365, and CLEVR for evaluation, while using the entire test set of SpatialMQA. All experiments are conducted on 4 NVIDIA A800 GPUs, with training typically taking 6–12 hours depending on the model size. To ensure robustness, each experiment is repeated three times with different random seeds, and we report the averaged results.

\noindent\textbf{Baselines.}
Our baselines include both general hallucination detection methods and recently proposed approaches specifically designed for OH detection.
For token-level logit-based methods, we consider Entropy~\cite{entropy}, Negative Log-Likelihood (NLL)~\cite{LURE}, and Internal Confidence~\cite{IC}. Among attention-based approaches, we include SVAR~\cite{SVAR}, which leverages attention scores to identify hallucinated objects.
For hidden-state-based methods, we adopt GLSIM~\cite{GLSIM} and ContextLens~\cite{Contextual_Lens}.

\noindent\textbf{Metrics.} 
AUROC (Area Under the Receiver Operating Characteristic Curve) and AUPR (Area Under the Precision–Recall Curve) are adopted as the primary detection metrics. 
To ensure that fine-tuning does not degrade the model’s original capability, we also report recall (for captioning) and accuracy (for QA). For our method, these results are obtained after fine-tuning, while the results of other methods correspond to the original models.


\subsection{Main Results.}
\begin{table*}[t]
\centering
\caption{Performance comparison on MSCOCO and Objects365 benchmarks. The best results are in bold, and the second best are underlined.}

\label{tab:main_results}
\setlength{\tabcolsep}{3.2pt}
\resizebox{\linewidth}{!}{
\begin{tabular}{l ccc ccc ccc ccc ccc}
\toprule

\multirow{3}{*}{\textbf{Method}}
& \multicolumn{15}{c}{\textbf{Models}} \\

\cmidrule(lr){2-16}
& 
\multicolumn{3}{c}{\textbf{LLaVA-1.5-7B}}
& \multicolumn{3}{c}{\textbf{LLaVA-1.5-13B}}
& \multicolumn{3}{c}{\textbf{LLaVA-NeXT-13B}}
& \multicolumn{3}{c}{\textbf{Qwen2.5-VL-7B}}
& \multicolumn{3}{c}{\textbf{LLaVA-OneVision1.5-8B}} \\

& 
Recall & AUROC & AUPR
& Recall & AUROC & AUPR
& Recall & AUROC & AUPR
& Recall & AUROC & AUPR
& Recall & AUROC & AUPR \\
\midrule


\multicolumn{16}{c}{\textbf{MSCOCO benchmark}} \\

NLL & 76.10 & 63.70 & 84.90
& 76.56 & 63.10 & 86.10
& 62.80 & 66.48 & 91.71
& 66.51 & 63.76 & 90.29
& 66.30 & 64.33 & 87.20 \\

Entropy & 76.10 & 64.00 & 85.00
& 76.56 & 63.20 & 86.30
& 62.80 & 66.01 & 91.63
& 66.51 & 66.11 & 90.78
& 66.30 & 65.56 & 86.79 \\

Internal Conf. & 76.10 & 72.90 & 89.30
& 76.56 & 71.00 & 90.00
& 62.80 & 75.12 & 95.05
& 66.51 & 63.91 & 91.56
& 66.30 & 71.89 & 93.07 \\

SVAR & 76.10 & 75.29 & 91.95
& 76.56 & 75.20 & 92.90
& 62.80 & 71.18 & 94.60
& 66.51 & 72.25 & 95.13
& 66.30 & 75.46 & 95.33 \\

Contextual Lens & 76.10 & 75.40 & 90.70
& 76.56 & 78.70 & 92.80
& 62.80 & 72.40 & 94.46
& 66.51 & 68.69 & 93.16
& 66.30 & 61.02 & 92.61 \\

GLSIM & 76.10 & \underline{83.70} & \underline{94.20}
& 76.56 & \underline{84.97} & \underline{95.15}
& 62.80 & \underline{78.37} & \underline{95.35}
& 66.51 & \underline{74.05} & \underline{94.40}
& 66.30 & \underline{76.94} & \underline{95.42} \\

\rowcolor{blue!15}
\textbf{TruthLens(Ours)} & 77.34 & \textbf{90.57} & \textbf{97.25}
& 77.30 & \textbf{90.99} & \textbf{97.59}
& 63.40 & \textbf{93.07} & \textbf{98.91}
& 66.80 & \textbf{91.46} & \textbf{98.59}
& 66.00 & \textbf{93.04} & \textbf{98.93} \\

\midrule


\multicolumn{16}{c}{\textbf{Objects365 benchmark}} \\

NLL & 28.23 & 62.90 & 60.80
& 29.12 & 59.40 & 61.00
& 26.40 & 52.99 & 65.80
& 32.33 & 59.19 & 75.11
& 28.65 & 57.95 & 71.99 \\

Entropy & 28.23 & 63.30 & 60.90
& 29.12 & 59.10 & 60.40
& 26.40 & 53.22 & 65.79
& 32.33 & 60.27 & 75.48
& 28.65 & 57.06 & 74.80 \\

Internal Conf. & 28.23 & 68.70 & 67.40
& 29.12 & 65.50 & 70.00
& 26.40 & \underline{71.74} & 78.32
& 32.33 & 61.35 & 77.10
& 28.65 & 64.10 & 79.95 \\

SVAR & 28.23 & 64.90 & 66.60
& 29.12 & 63.50 & 68.20
& 26.40 & 70.33 & \underline{79.02}
& 32.33 & \underline{65.11} & \underline{80.75}
& 28.65 & \underline{72.59} & \underline{80.88} \\

Contextual Lens & 28.23 & 63.20 & 62.60
& 29.12 & 62.10 & 65.60
& 26.40 & 65.81 & 75.82
& 32.33 & 62.57 & 79.72
& 28.65 & 55.31 & 68.70 \\

GLSIM & 28.23 & \underline{72.60} & \underline{74.60}
& 29.12 & \underline{70.40} & \underline{74.00}
& 26.40 & 69.76 & 80.07
& 32.33 & 62.36 & 79.70
& 28.65 & 65.30 & 75.30 \\

\rowcolor{blue!15}
\textbf{TruthLens(Ours)} & 27.90 & \textbf{77.89} & \textbf{84.56}
& 29.20 & \textbf{77.69} & \textbf{85.36}
& 26.70 & \textbf{80.15} & \textbf{87.58}
& 32.30 & \textbf{71.54} & \textbf{84.56}
& 28.70 & \textbf{82.14} & \textbf{88.31} \\

\bottomrule
\end{tabular}
}
\end{table*}
\begin{table*}[t]
\centering
\caption{Performance comparison on CLEVR and SpatialMQA benchmarks. Appendix Table.~\ref{tab:attribute_results_with_std} reports
the results with standard deviations.}

\label{tab:attribute_results}
\setlength{\tabcolsep}{3.2pt}

\resizebox{\linewidth}{!}{
\begin{tabular}{l ccc ccc ccc ccc ccc}
\toprule

\multirow{3}{*}{\textbf{Method}}
& \multicolumn{15}{c}{\textbf{Models}} \\

\cmidrule(lr){2-16}
& 
\multicolumn{3}{c}{\textbf{LLaVA-1.5-7B}}
& \multicolumn{3}{c}{\textbf{LLaVA-1.5-13B}}
& \multicolumn{3}{c}{\textbf{LLaVA-NeXT-13B}}
& \multicolumn{3}{c}{\textbf{Qwen2.5-VL-7B}}
& \multicolumn{3}{c}{\textbf{LLaVA-OneVision1.5-8B}} \\

& 
ACC & AUROC & AUPR
& ACC & AUROC & AUPR
& ACC & AUROC & AUPR
& ACC & AUROC & AUPR
& ACC & AUROC & AUPR \\
\midrule

\multicolumn{16}{c}{\textbf{CLEVR benchmark}} \\

NLL & 43.43 & 45.22 & 38.13
& 47.03 & 44.42 & 41.30
& 49.70 & 52.56 & 49.44
& 97.70 & \underline{88.72} & \underline{99.64}
& 97.98 & 89.39 & 99.74 \\

Entropy & 43.43 & 44.97 & 38.02
& 47.03 & 51.92 & 48.21
& 49.70 & 51.71 & 49.11
& 97.70 & 88.54 & 99.63
& 97.98 & \underline{90.29} & \underline{99.76} \\

Internal Conf. & 43.43 & 37.78 & 35.67
& 47.03 & 45.18 & 47.25
& 49.70 & 42.79 & 44.01
& 97.70 & 32.77 & 95.24
& 97.98 & 35.98 & 96.68 \\

SVAR & 43.43 & 55.39 & 49.13
& 47.03 & 50.36 & 50.12
& 49.70 & 42.79 & 44.01
& 97.70 & 73.69 & 97.31
& 97.98 & 75.75 & 99.30 \\

Contextual Lens & 43.43 & 57.16 & 49.28
& 47.03 & \underline{58.03} & 52.77
& 49.70 & 58.08 & 57.97
& 97.70 & 82.37 & 99.47
& 97.98 & 71.45 & 98.94 \\

GLSIM & 43.43 & \underline{59.89} & \underline{50.94}
& 47.03 & 57.96 & \underline{54.45}
& 49.70 & \underline{60.45} & \underline{61.42}
& 97.70 & 79.72 & 99.26
& 97.98 & 82,60 & 99.31 \\

\rowcolor{blue!15}
\textbf{TruthLens(Ours)} & 43.73 & \textbf{63.34} & \textbf{53.85}
& 47.15 & \textbf{60.87} & \textbf{54.57}
& 48.52 & \textbf{63.79} & \textbf{69.76}
& 97.96 & \textbf{92.11} & \textbf{99.77}
& 97.92 & \textbf{92.33} & \textbf{99.78} \\

\midrule

\multicolumn{16}{c}{\textbf{SpatialMQA benchmark}} \\

NLL & 30.41 & \underline{62.42} & 40.14
& 32.69 & 57.76 & 41.89
& 34.65 & 60.46 & 46.61
& 36.90 & 59.88 & 51.92
& 35.73 & 60.40 & 48.09 \\

Entropy & 30.41 & 62.33 & \underline{41.84}
& 32.69 & 51.73 & 36.12
& 34.65 & 56.68 & 46.07
& 36.90 & 61.72 & \underline{52.86}
& 35.73 & 60.23 & 48.13 \\

Internal Conf. & 30.41 & 46.41 & 28.53
& 32.69 & 51.85 & 36.40
& 34.65 & 56.68 & 46.07
& 36.90 & 51.35 & 37.16
& 35.73 & 51.65 & 38.54 \\

SVAR & 30.41 & 56.40 & 36.07
& 32.69 & \underline{66.61} & \underline{50.32}
& 34.65 & 65.02 & 50.36
& 36.90 & 61.27 & 48.30
& 35.73 & 57.00 & 44.19 \\

Contextual Lens & 30.41 & 62.36 & 41.27
& 32.69 & 62.44 & 45.58
& 34.65 & 64.87 & 52.75
& 36.90 & \underline{64.84} & 52.44
& 35.73 & 61.03 & 48.41 \\

GLSIM & 30.41 & 61.30 & 40.69
& 32.69 & 64.89 & \textbf{51.11}
& 34.65 & \underline{66.49} & \underline{57.55}
& 36.90 & 62.49 & 52.03
& 35.73 & \underline{63.99} & \underline{49.13} \\

\rowcolor{blue!15}
\textbf{TruthLens(Ours)} & 31.17 & \textbf{71.32} & \textbf{49.71}
& 32.76 & \textbf{67.84} & 47.93
& 34.52 & \textbf{67.05} & \textbf{57.94}
& 36.92 & \textbf{76.06} & \textbf{68.80}
& 36.48 & \textbf{76.04} & \textbf{66.60} \\

\bottomrule
\end{tabular}
}
\end{table*}
As shown in Table.~\ref{tab:main_results}, our method achieves the best performance on both MSCOCO and Object365 across all evaluated LVLMs, while preserving the models’ original task performance.
On MSCOCO, our approach consistently outperforms existing baselines across different architectures. For example, on LLaVA-OneVision-1.5-8B, it achieves an AUROC of 93.04 and an AUPR of 98.93, improving over the strongest baseline (GLSIM) by 16.10\% in AUROC and 3.51\% in AUPR. Similar improvements are also observed on other models.
On the more challenging Object365 benchmark, which involves a larger object class, our method maintains strong performance. On LLaVA-OneVision-1.5-8B, it surpasses the strongest competing method (SVAR) by 9.55\% in AUROC and 7.43\% in AUPR, with consistent gains across other models.
Notably, although the model is fine-tuned only on MSCOCO, it generalizes to Object365 without additional training. This cross-dataset transfer suggests that the method captures intrinsic truthfulness-related signals rather than dataset-specific object statistics.

\noindent\textbf{Results on OAH detection benchmarks.} 
As shown in Table.~\ref{tab:attribute_results}, our method achieves competitive performance on both CLEVR and SpatialMQA for OAH detection.
Models with weaker fine-grained attribute reasoning ability (e.g., LLaVA-1.5) generally obtain lower OAH detection performance across all methods. This likely reflects limited representational capacity, resulting in reduced separability between real and hallucinated token representations in the feature space. Nevertheless, our method remains consistently competitive even under these conditions.
On stronger models such as Qwen2.5-VL and LLaVA-OneVision-1.5, the gains become more substantial. For example, on Qwen2.5-VL-7B under SpatialMQA, AUROC improves from 64.84 (Contextual Lens) to 76.06, an absolute gain of 11.22\%. Similar trends are observed on CLEVR.
Overall, these results suggest that our approach generalizes well to OAH detection and benefits from stronger multimodal representations.

\subsection{Further Analysis}
\label{further_analysis}
For consistency, MSCOCO is used as the test set in all subsequent experiments.

\noindent\textbf{Ablation study.}
We conduct ablation experiments to evaluate the contributions of the MSE loss, loss reweighting, and KL divergence constraint. As shown in Table~\ref{tab:ablation}, directly applying the proposed truthfulness score to the original pretrained model without fine-tuning yields poor OH detection performance, e.g., 34.80 AUROC on LLaVA-1.5-7B. Introducing the MSE objective with the KL constraint substantially improves performance, increasing AUROC to 89.02 on LLaVA-1.5-7B and 85.69 on Qwen2.5-VL-7B, indicating that aligning token-level truthfulness scores with target rewards is essential. Further adding loss reweighting achieves the best results across models, reaching 90.57 and 91.46 AUROC on LLaVA-1.5-7B and Qwen2.5-VL-7B, respectively. This gain comes from mitigating the imbalance between real and hallucinated tokens, which is especially severe for stronger models such as Qwen2.5-VL. Finally, removing the KL constraint causes the model to deviate significantly from the original pretrained model and severely degrades its generative ability, sometimes producing repetitive captions regardless of the input image. These results confirm that KL regularization is crucial for preserving the model’s original functionality while enabling effective hallucination detection.

\noindent\textbf{Impact of Special Token Selection.} The vocabularies of LVLMs contain multiple special tokens that are rarely generated during autoregressive decoding. In our main experiments, we adopt <|image\_pad|> as the designated special token for Qwen2.5-VL and LLaVA-OneVision-1.5. To examine whether our method depends on a particular token choice, we further evaluate alternative special tokens, including <|vision\_start|> and <|vision\_end|>.
As shown in Table.~\ref{tab:special_token}, the performance differences across different special tokens are marginal. For example, on Qwen2.5-VL-7B, AUROC varies only within a narrow range (91.20–91.54), and similar stability is observed on LLaVA-OneVision-1.5-8B. These results indicate that our method is not sensitive to the specific selection of special tokens.
This robustness suggests that the effectiveness of our approach stems from the underlying mechanism of aligning hidden representations with the truthfulness signal, rather than from any particular property of a specific special token. Consequently, our framework is flexible and can be readily adapted to LVLMs with different vocabularies and special-token designs.
\begin{table}[t]
\centering
\captionsetup{font=small, labelfont=bf}
\small

\begin{minipage}{0.48\linewidth}
\centering

\caption{Ablation results on the effectiveness of each component
in TruthLens.}

\label{tab:ablation}
\resizebox{\linewidth}{!}{%
\begin{tabular}{ccc cc cc}
\toprule
$L_{\text{MSE}}$ & $KL$ & Reweight
& \multicolumn{2}{c}{LLaVA-1.5-7B}
& \multicolumn{2}{c}{Qwen2.5-VL-7B} \\
\cmidrule(lr){4-5} \cmidrule(lr){6-7}
 &  &  & AUROC$\uparrow$ & AUPR$\uparrow$
      & AUROC$\uparrow$ & AUPR$\uparrow$ \\
\midrule
-          & -          & -          & 34.80 & 70.71 & 35.04    & 80.98    \\
\checkmark & \checkmark & -          & 89.02    & 96.36    & 85.69    & 97.49    \\
\checkmark & \checkmark & \checkmark & \textbf{90.57}    & \textbf{97.25}   & \textbf{91.46}    & \textbf{98.59}    \\
\bottomrule
\end{tabular}
}
\end{minipage}
\hfill 
\begin{minipage}{0.48\linewidth}
\centering

\caption{Performance comparison between different special tokens.}

\label{tab:special_token}
\resizebox{\linewidth}{!}{%
\begin{tabular}{c cc cc}
\toprule
\multirow{2}{*}{Special token}
& \multicolumn{2}{c}{Qwen2.5-VL-7B}
& \multicolumn{2}{c}{LLaVA-OneVision-1.5-8B} \\
\cmidrule(lr){2-3} \cmidrule(lr){4-5}
& AUROC$\uparrow$ & AUPR$\uparrow$
& AUROC$\uparrow$ & AUPR$\uparrow$ \\
\midrule
$\text{<|image\_pad|>}$ & 91.46 & 98.59 & 93.04 & 98.93 \\
$\text{<|vision\_start|>}$ & \textbf{91.54} & \textbf{98.64} & 92.87 & 98.90 \\
$\text{<|vision\_end|>}$ & 91.20 & 98.52 & \textbf{93.48} & \textbf{99.02} \\
\bottomrule
\end{tabular}%
}
\end{minipage}

\end{table}

\noindent\textbf{General capability preservation.}
To further examine whether TruthLens affects the general multimodal ability of the base LVLM, we evaluate the fine-tuned LLaVA1.5-7B on 13 widely used general benchmarks. As shown in Table~\ref{tab:general_benchmarks}, the performance after fine-tuning remains nearly unchanged across all benchmarks. Specifically, the fine-tuned model obtains comparable results on VQAv2\cite{VQAv2}, GQA\cite{GQA}, SQA-IMG\cite{SQA}, POPE\cite{pope}, MMBench\cite{MMBench}, LLaVABench\cite{llava}, and RealWorldQA\cite{xai2024realworldqa}, with only negligible fluctuations. Meanwhile, it slightly improves on several benchmarks such as VizWiz\cite{Vizwiz}, TextVQA\cite{TextVQA}, MME\cite{MME}, MMBenchCN\cite{MMBench}, SEED-Bench\cite{SEEDBench}, and MM-Vet\cite{MMVet}. These results indicate that TruthLens does not sacrifice the original general-purpose multimodal capability of LLaVA1.5-7B. We attribute this stability to the KL regularization, which constrains the fine-tuned model to preserve the generation behavior of the original LVLM while learning hallucination-aware signals.

\begin{table}[t]
\centering
\caption{General benchmark results of Raw and fine-tuned LLaVA1.5-7B.}
\label{tab:general_benchmarks}
\scriptsize
\begin{tabular}{lccc@{\hspace{0.8em}}lccc}
\toprule
Benchmark & Raw & FT & $\Delta$ 
& Benchmark & Raw & FT & $\Delta$ \\
\midrule
VQAv2\cite{VQAv2}       & 75.59 & 75.65 & +0.06 
& MMBench\cite{MMBench}     & 64.08 & 64.08 & +0.00 \\

GQA\cite{GQA}         & 68.46 & 68.51 & +0.05
& MMBenchCN\cite{MMBench}   & 54.84 & 55.07 & +0.23 \\

VizWiz\cite{Vizwiz}      & 53.68 & 54.16 & +0.48
& SEED-Bench\cite{SEEDBench}  & 59.03 & 59.21 & +0.18 \\

SQA-IMG\cite{SQA}     & 65.24 & 65.19 & -0.05
& LLaVABench\cite{llava}  & 63.40 & 63.20 & -0.20 \\

TextVQA\cite{TextVQA}     & 57.65 & 57.94 & +0.29
& MM-Vet\cite{MMVet}      & 30.50 & 30.80 & +0.30 \\

POPE\cite{pope}        & 85.76 & 85.67 & -0.09
& RealWorldQA\cite{xai2024realworldqa} & 54.24 & 54.03 & -0.21 \\

MME\cite{MME}         & 1468.56 & 1474.59 & +6.03
&             &       &       &       \\
\bottomrule
\end{tabular}
\end{table}

\begin{figure}[t]
    \centering
    \includegraphics[width=\linewidth]{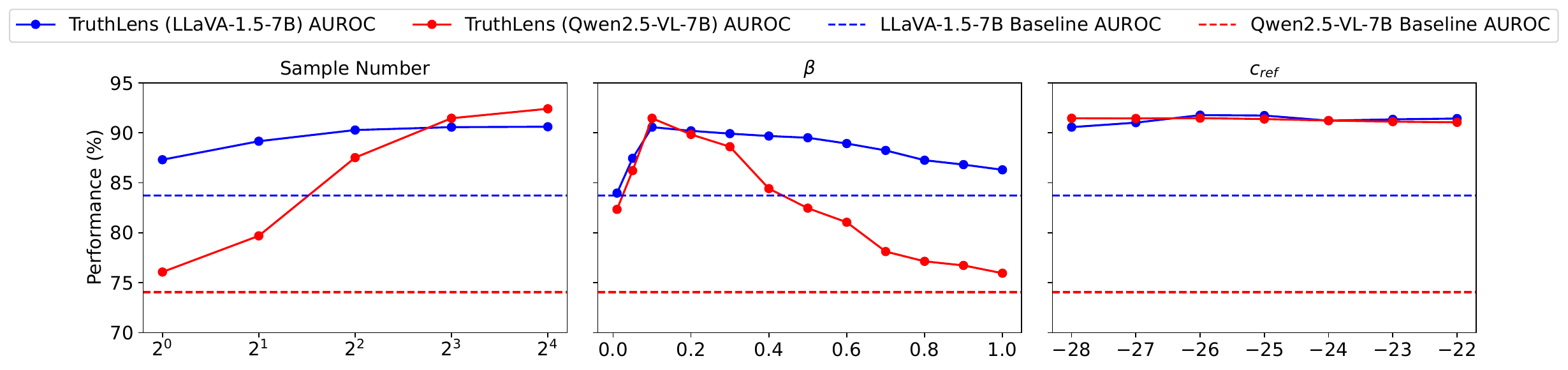}

    \caption{Sensitivity studies of hyperparameters on LLaVA-1.5-7B and Qwen2.5-VL-7B. Dashed lines indicate the strongest baselines.}
    \label{fig:sensitivity}

\end{figure}

\noindent\textbf{Hyperparameter Sensitivity Analysis.}
As shown in Fig.~\ref{fig:sensitivity}, we performed a series of hyperparameter sensitivity analyses on LLaVA-1.5-7B and Qwen2.5-VL-7B to investigate the effects of key design choices.


We first analyze the effect of the coefficient $\beta$, which controls the scale of the token-level truthfulness score. According to Eq.~\ref{reweight_loss}, the gradient satisfies
$\frac{\partial \mathcal{L}}{\partial \log \pi_\theta(z_c \mid x, y_{<o})} \propto \beta$,
indicating that $\beta$ directly determines the strength of the reward regression signal.
As shown in Fig.~\ref{fig:sensitivity} (middle), a large $\beta$ ($\beta > 1.0$) amplifies the variance of token probabilities, causing overly dispersed truthfulness scores and unstable optimization. In contrast, a very small $\beta$ ($\beta < 0.01$) compresses the reward scale, leading to conservative updates and slower convergence. Therefore, we adopt a moderate-to-small $\beta$ to balance optimization stability and effective reward regression.


We next investigate the number of sampled responses during training. As shown in Fig.~\ref{fig:sensitivity} (left), performance consistently improves as the number of rollouts increases, indicating that multi-sample estimation reduces variance and enhances robustness. However, the gains gradually saturate beyond $2^3$ rollouts, suggesting diminishing returns from additional sampling. Notably, our method consistently surpasses the current SOTA (dashed lines) under all rollout settings—even with a single rollout—demonstrating that the improvements primarily stem from our reward modeling design rather than increased sampling. This highlights the strong sample efficiency of our approach.


Finally, we analyze the influence of the reference constant $c_{\text{ref}}$, which serves as a baseline shift in the truthfulness score. Since the log-probabilities of special tokens are typically very small (often below -20), $c_{\text{ref}}$ must be chosen at a comparable scale to ensure stable optimization. As shown in Fig.~\ref{fig:sensitivity} (right), performance remains stable within a moderately negative range of $c_{\text{ref}}$ and consistently exceeds the SOTA baseline. However, if $c_{\text{ref}}$ is set too large (i.e., less negative), the induced score shift becomes excessive, distorting the original log-probability landscape and interfering with the model’s language modeling behavior. In such cases, the regression objective may dominate optimization, leading to degraded performance. Overall, our method demonstrates stable behavior across a wide range of hyperparameters while consistently outperforming the SOTA baseline.

\begin{figure}[t]
\centering
\small

\begin{minipage}{0.48\linewidth}
\centering
\captionsetup{font=small, labelfont=bf, type=table} 
\setlength{\tabcolsep}{4.5pt}

\caption{Performance comparison between different sub-tokens.}
\label{tab:multi_token}
\resizebox{\linewidth}{!}{%
\begin{tabular}{c cc cc}
\toprule
\multirow{2}{*}{Token}
& \multicolumn{2}{c}{Qwen2.5-VL-7B}
& \multicolumn{2}{c}{LLaVA-OneVision-1.5-8B} \\
\cmidrule(lr){2-3} \cmidrule(lr){4-5}
& AUROC$\uparrow$ & AUPR$\uparrow$
& AUROC$\uparrow$ & AUPR$\uparrow$ \\
\midrule
First & 91.46 & \textbf{98.59} & 93.04 & 98.93 \\
Last & 91.39 & 98.57 & \textbf{93.14} & \textbf{98.97} \\
Average & \textbf{91.73} & 98.50 & 92.07 & 98.68 \\
\bottomrule
\end{tabular}%
}
\end{minipage}
\hfill
\begin{minipage}{0.48\linewidth}
\centering
\captionsetup{font=small, labelfont=bf, type=figure} 
\includegraphics[width=\linewidth]{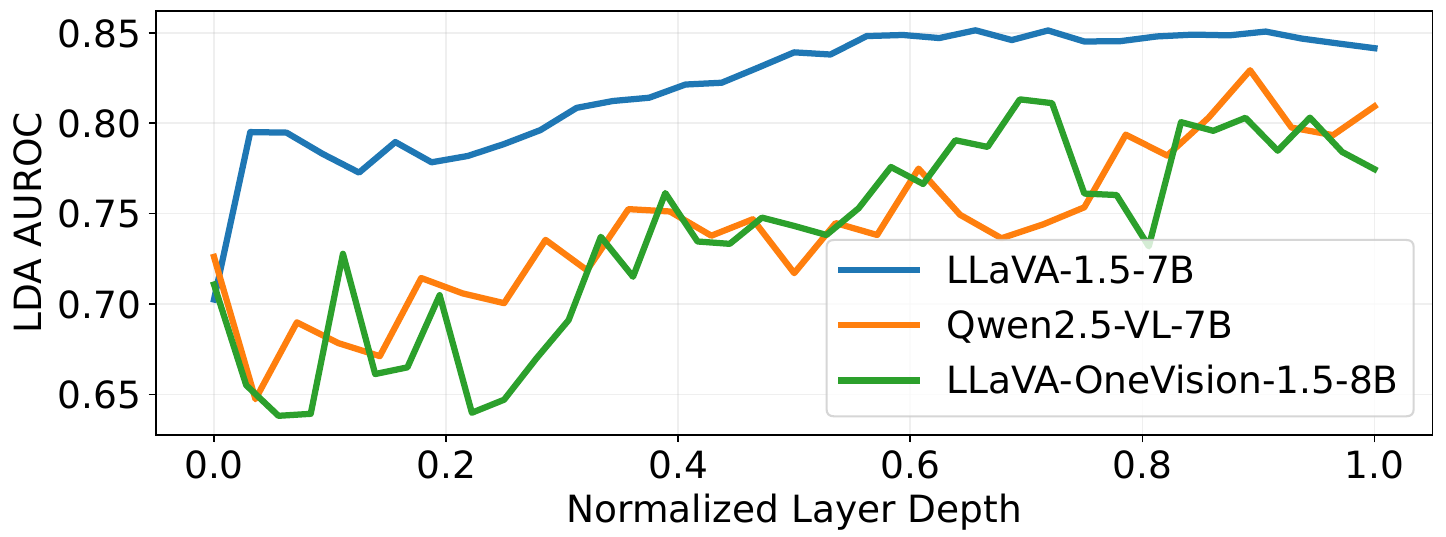}

\caption{Evolution of token separability.}
\label{fig:layer_wise_separability}
\end{minipage}

\end{figure}

\noindent\textbf{Multi-token Objects.} When an object mention is tokenized into multiple sub-tokens, we apply the loss to the first token by default. Nevertheless, alternative strategies are also possible, including imposing the constraint on the last token or averaging the truthfulness scores across all sub-tokens corresponding to the object.
As shown in Table.~\ref{tab:multi_token}, these different strategies yield highly comparable performance. The differences between applying the loss to the first token, the last token, and averaging across sub-tokens are marginal across both evaluated models. This consistency indicates that the truthfulness signal is not confined to a specific sub-token position.
Instead, the separability between real and hallucinated object representations appears to be distributed throughout the token-level hidden states of the object. Such robustness suggests that our method does not depend on a carefully selected token position and remains stable under different supervision placements. For simplicity and implementation convenience, we therefore impose the constraint on the first token of each object.

\section{Further Discussion}

\textbf{Emergence of token separability.}
In Section~\ref{motivation}, we showed that real and hallucinated object tokens are clearly separable in the last-layer hidden representations. We further investigate how this separability emerges across the network by extracting hidden states from each Transformer layer and applying LDA independently.
As shown in Fig.~\ref{fig:layer_wise_separability}, token separability consistently increases with depth across all three LVLMs. The most pronounced improvement appears in the middle layers (normalized depth 0.3--0.7), where the LDA AUROC rises steadily before plateauing in later layers. This trend suggests that the distinction between real and hallucinated tokens is gradually formed rather than abruptly emerging at the final layer.
Recent studies~\cite{SVAR,zhang2025cross,fan2025mathcal} show that intermediate LVLM layers are responsible for intensive cross-modal interaction. The sharp separability improvement in these layers indicates that visual grounding is critical for distinguishing real from hallucinated tokens: grounded object tokens become increasingly aligned with image features, whereas hallucinated tokens, mainly driven by language priors, fail to establish such alignment. This progressively enlarges their representational divergence, resulting in stronger separability in deeper layers.

\noindent \textbf{LM head failure in token discrimination.}
To understand why the separability observed in hidden representations does not translate into discriminative LM head outputs, we analyze whether token truthfulness-related information can be effectively read out by the LM head weight matrix $\mathbf{W}$. Let $\mathbf{v}=\mathbf{m}_r-\mathbf{m}_h$ denote the difference between the mean hidden states of real and hallucinated tokens. Following the output-null framework~\cite{Cortical_activity_in_the_null_space}, a discriminative direction can affect downstream outputs only if it lies in the effective subspace of the readout matrix; otherwise, it may remain functionally silent even when separable in representation space. We therefore quantify the alignment between $\mathbf{v}$ and the LM head projection space as:
\begin{equation}
\small
\text{Ratio} = \frac{|\mathbf{Wv}|}{|\mathbf{W}|_F , |\mathbf{v}|},
\end{equation}
where $|\cdot|_F$ denotes the Frobenius norm. This ratio measures the normalized projection strength of the discriminative direction and is bounded in $(0,1]$.

We conduct this analysis on MSCOCO using 5{,}000 sampled images for each model. The resulting ratios are 0.0237, 0.0203, and 0.0154 for LLaVA-1.5-7B, Qwen2.5-VL-7B, and LLaVA-OneVision-1.5-8B, respectively. These consistently small values indicate that the truthfulness-related direction is only weakly aligned with the LM head projection space across different LVLM architectures. Thus, although real and hallucinated tokens are clearly separable in feature space, this information is not effectively exposed by the vocabulary projection, revealing a systematic representation--projection mismatch.

\noindent\textbf{Detection-to-Mitigation.}
Beyond hallucination detection, we further investigate whether TruthLens can provide useful signals for mitigating hallucinated content in LVLM generations. Specifically, we adopt a simple yet effective revision strategy to close the detection-to-mitigation loop. Given an image caption generated by LLaVA-1.5-7B, we first apply TruthLens to score the generated tokens and identify potential hallucinations according to the threshold selected under the 95\% TPR setting. Tokens whose TruthLens scores fall below the threshold are regarded as potentially hallucinated content. We then feed both the original caption and the detected hallucinated tokens back to the captioning model, prompting it to revise the caption by removing unsupported or visually inconsistent content while preserving the correct descriptions.

We conduct this experiment on 5,000 MSCOCO images with LLaVA-1.5-7B. The results show that the proposed revision strategy effectively reduces hallucinations in the generated captions. Specifically, $\mathrm{CHAIR}_i$ decreases from 14.81 to 12.23, and $\mathrm{CHAIR}_s$ decreases from 49.40 to 43.78, indicating that both instance-level and sentence-level hallucinations are reduced after revision. Meanwhile, the recall score only slightly changes from 76.74 to 76.31, suggesting that the revision process does not simply remove a large number of object descriptions, but instead selectively suppresses hallucinated content while largely preserving valid visual information. These results demonstrate that TruthLens is not only effective for detecting hallucinated tokens, but can also serve as a practical guidance signal for hallucination mitigation, enabling a lightweight detection-to-mitigation pipeline without requiring external detectors or retraining of the captioning model.
\section{Conclusion}
This study reveals that hidden representations of real and hallucinated tokens in LVLMs exhibit clear separability in feature space. However, this separability is not reflected in the predictions of the LM head. To address this representation–projection mismatch, we fine-tune the model to explicitly enable the LM head to read out the underlying discriminative signal.
Specifically, we define the truthfulness score of an object token as the difference between the log-probability assigned by the model to a designated special token at the corresponding token position and a predefined constant. During training, we introduce an MSE loss to align real and hallucinated tokens with their respective reward signals. At inference time, the truthfulness score can be directly used to determine whether a token is hallucinated. Extensive experiments across multiple LVLM architectures and diverse benchmarks demonstrate the effectiveness and robustness of our approach. In future work, we aim to explore how this mechanism can be further extended to mitigate hallucinations in LVLMs more directly, beyond detection toward generation-time correction.

 \noindent\textbf{Acknowledgments}. This work is supported in part by the National Natural Science Foundation of China (grant No. 62571559), the Major Key Project of PCL (grant No. PCL2025AS209), and Guangdong Excellent Youth Team Program (grant No. 2023B1515040025).

\bibliographystyle{splncs04}
\bibliography{main}
\clearpage
\appendix
\section{Appendix}
\subsection{Baselines}
\label{baselines_intro}
\textbf{Negative Log-likelihood}~\cite{LURE}.
To quantify the hallucination risk in autoregressive object generation, Zhou et al proposes the negative log-likelihood framework.
In this setup, the probability of generating an object token $o$ at position $j$ is modeled as $p(o \mid \mathbf{y}_{<j}, \mathbf{v})$, where $\mathbf{y}_{<j}$ denotes the sequence of previously generated tokens and $\mathbf{v}$ represents the visual context.
The hallucination score for each object $o$ is derived from this probability:
\begin{equation}
\label{eq:nll}
s_{\text{nll}} = -\log p(o \mid \mathbf{y}_{<j}, \mathbf{v}).
\end{equation}
To ensure consistency with the score definition established in Eq.~\ref{eq:nll}, we redefine this score as $s'_{\text{nll}} = -s_{\text{nll}}$.

\noindent\textbf{Entropy}~\cite{entropy}.
To further characterize object-level hallucination, we compute an entropy-based score by estimating the uncertainty of the token probability distribution at position \(j\). This is done by calculating the entropy of the distribution \(p(y \mid \mathbf{y}_{<j}, \mathbf{v})\) over the vocabulary \(\mathcal{V}\):
\begin{equation}
s_{\text{entropy}} = -\sum_{y \in \mathcal{V}} p(y \mid \mathbf{y}_{<j}, \mathbf{v}) \log p(y \mid \mathbf{y}_{<j}, \mathbf{v}).
\end{equation}
To maintain consistency with the score definition in Equation (1), we use the negated form \(s'_{\text{entropy}} = -s_{\text{entropy}}\).

\noindent\textbf{Internal Confidence}~\cite{IC}.
To measure the model's internal confidence in its predictions, we draw on the logit lens framework introduced by Jiang et al~\cite{IC}, which allows us to trace how visual features are mapped to textual outputs across different layers of the network. For the task of object hallucination detection, we define the \textbf{internal confidence score} as the maximum softmax probability assigned to the object word \(o\) across all image representations and network layers. The hallucination score is computed as:
\begin{equation}
s_{\text{IC}} = \max_{l \in [L]} \max_{i \in [N]} \text{VLL}_l(v_i)[o],
\end{equation}
where \(L\) is the total number of layers in the model, and \(N\) is the total number of image patches.

\noindent\textbf{Summed Visual Attention Ratio (SVAR)}~\cite{SVAR}.
The Visual Attention Ratio (VAR) is designed to quantify the degree to which a generated token \(o\) grounds its prediction in visual information. It is calculated by summing the attention weights that token \(o\) assigns to all image tokens \(v_i\) within a specific attention head \(h\) and network layer \(\ell\):
\begin{equation}
\text{VAR}^{(\ell,h)}(o) \triangleq \sum_{i=1}^{N} A^{(\ell,h)}(o, v_i),
\end{equation}
where \(A^{(\ell,h)}(o, v_i)\) denotes the attention weight from object token \(o\) to image token \(v_i\) in the \(h\)-th attention head of the \(\ell\)-th layer.

Building on this foundational measure, we adopt the Summed Visual Attention Ratio (SVAR) proposed by Jiang et al. \cite{SVAR}. This metric aggregates the VAR scores to provide a holistic view of the model's visual grounding. Specifically, it averages the VAR across all attention heads and sums the results over a predefined range of layers. For an object token \(o\), we compute the SVAR score by considering the model's behavior from layer \(\ell_5\) to \(\ell_{18}\):
\begin{equation}
s_{\text{SVAR}} = \frac{1}{H} \sum_{\ell=5}^{18} \sum_{h=1}^{H} \text{VAR}^{(\ell,h)}(o),
\end{equation}
where \(H\) represents the total number of attention heads in the model.

\noindent\textbf{GLSIM}~\cite{GLSIM} is designed to quantify the likelihood that a generated object token $o$ is grounded in the input image by measuring both scene-level semantic compatibility and region-level visual evidence. Specifically, it combines two complementary signals: a global similarity score and a local similarity score computed from the latent representations of the LVLM.

The local similarity evaluates whether the object token is visually supported by specific image regions. Given the set of visual tokens $v=\{v_1,\dots,v_N\}$, we first identify the Top-$K$ image tokens that are most relevant to object $o$, denoted as $I(o)$. The local similarity is computed as the average cosine similarity between the object embedding and the embeddings of these selected visual tokens:
\begin{equation}
s_{\text{local}}(o,x)=\frac{1}{K}\sum_{v_i\in I(o)} \mathrm{sim}(h_l(v_i),h_{l'}(o)),
\end{equation}
where $\mathrm{sim}(\cdot,\cdot)$ denotes cosine similarity.

The global similarity measures the semantic consistency between the object token and the overall scene representation. Specifically, it computes the cosine similarity between the object embedding and the hidden representation of the final instruction token:
\begin{equation}
s_{\text{global}}(o,x)=\mathrm{sim}\big(h_l(v,t),h_{l'}(o)\big).
\end{equation}

Finally, the GLSIM score combines these two complementary signals:
\begin{equation}
s_{\text{GLSIM}}(o,x)=w\,s_{\text{global}}(o,x)+(1-w)\,s_{\text{local}}(o,x),
\end{equation}
where $w\in[0,1]$ balances global scene-level semantics and local visual grounding.

\subsection{\textbf{Additional Experiments and Analysis}}
\textbf{Results on MSCOCO, Object365, CLEVR and SpatialMQA.}
As shown in Table.~\ref{tab:OH_results_with_std} and~\ref{tab:attribute_results_with_std}, we report the detailed performance of each method, with the corresponding standard deviations provided to demonstrate performance stability. Since SpatialMQA is evaluated on the entire test set, the baseline methods do not report variance.

\begin{table}[h]
\centering
\caption{Performance comparison on MSCOCO and Objects365 benchmarks. The best results are in bold, and the second best are underlined.}
\label{tab:OH_results_with_std}
\setlength{\tabcolsep}{3.2pt}
\resizebox{\linewidth}{!}{
\begin{tabular}{l ccc ccc ccc ccc ccc}
\toprule

\multirow{3}{*}{\textbf{Method}}
& \multicolumn{15}{c}{\textbf{Models}} \\

\cmidrule(lr){2-16}
& 
\multicolumn{3}{c}{\textbf{LLaVA-1.5-7B}}
& \multicolumn{3}{c}{\textbf{LLaVA-1.5-13B}}
& \multicolumn{3}{c}{\textbf{LLaVA-NeXT-13B}}
& \multicolumn{3}{c}{\textbf{Qwen2.5-VL-7B}}
& \multicolumn{3}{c}{\textbf{LLaVA-OneVision1.5-8B}} \\

& 
Recall & AUROC & AUPR
& Recall & AUROC & AUPR
& Recall & AUROC & AUPR
& Recall & AUROC & AUPR
& Recall & AUROC & AUPR \\
\midrule

\multicolumn{16}{c}{\textbf{MSCOCO benchmark}} \\

NLL & $76.10_{\pm0.57}$ & $63.70_{\pm1.42}$ & $84.90_{\pm1.11}$
& $76.56_{\pm0.53}$ & $63.10_{\pm1.21}$ & $86.10_{\pm1.52}$
& $62.80_{\pm0.74}$ & $66.48_{\pm1.67}$ & $91.71_{\pm1.96}$
& $66.51_{\pm1.81}$ & $63.76_{\pm1.22}$ & $90.29_{\pm0.92}$
& $66.30_{\pm0.34}$ & $64.33_{\pm2.63}$ & $87.20_{\pm1.84}$ \\

Entropy & $76.10_{\pm0.57}$ & $64.00_{\pm1.21}$ & $85.00_{\pm1.66}$
& $76.56_{\pm0.53}$ & $63.20_{\pm2.04}$ & $86.30_{\pm1.59}$
& $62.80_{\pm0.74}$ & $66.01_{\pm2.26}$ & $91.63_{\pm1.34}$
& $66.51_{\pm1.81}$ & $66.11_{\pm1.24}$ & $90.78_{\pm0.89}$
& $66.30_{\pm0.34}$ & $65.56_{\pm1.89}$ & $86.79_{\pm1.37}$ \\

Internal Conf. & $76.10_{\pm0.57}$ & $72.90_{\pm1.35}$ & $89.30_{\pm1.14}$
& $76.56_{\pm0.53}$ & $71.00_{\pm2.46}$ & $90.00_{\pm1.99}$
& $62.80_{\pm0.74}$ & $75.12_{\pm3.13}$ & $95.05_{\pm1.88}$
& $66.51_{\pm1.81}$ & $63.91_{\pm3.71}$ & $91.56_{\pm2.53}$
& $66.30_{\pm0.34}$ & $71.89_{\pm1.76}$ & $93.07_{\pm1.69}$ \\

SVAR & $76.10_{\pm0.57}$ & $75.29_{\pm1.44}$ & $91.95_{\pm1.28}$
& $76.56_{\pm0.53}$ & $75.20_{\pm1.42}$ & $92.90_{\pm1.11}$
& $62.80_{\pm0.74}$ & $71.18_{\pm1.04}$ & $94.60_{\pm0.83}$
& $66.51_{\pm1.81}$ & $72.25_{\pm1.09}$ & $95.13_{\pm0.49}$
& $66.30_{\pm0.34}$ & $75.46_{\pm1.69}$ & $95.33_{\pm1.31}$ \\

Contextual Lens & $76.10_{\pm0.57}$ & $75.40_{\pm2.24}$ & $90.70_{\pm1.95}$
& $76.56_{\pm0.53}$ & $78.70_{\pm1.18}$ & $92.80_{\pm0.78}$
& $62.80_{\pm0.74}$ & $72.40_{\pm0.97}$ & $94.46_{\pm0.53}$
& $66.51_{\pm1.81}$ & $68.69_{\pm1.23}$ & $93.16_{\pm0.34}$
& $66.30_{\pm0.34}$ & $61.02_{\pm2.36}$ & $92.61_{\pm1.63}$ \\

GLSIM & $76.10_{\pm0.57}$ & $\underline{83.70}_{\pm1.52}$ & $\underline{94.20}_{\pm1.31}$
& $76.56_{\pm0.53}$ & $\underline{84.97}_{\pm0.42}$ & $\underline{95.15}_{\pm0.66}$
& $62.80_{\pm0.74}$ & $\underline{78.37}_{\pm1.32}$ & $\underline{95.35}_{\pm1.07}$
& $66.51_{\pm1.81}$ & $\underline{74.05}_{\pm1.57}$ & $\underline{94.40}_{\pm0.33}$
& $66.30_{\pm0.34}$ & $\underline{76.94}_{\pm2.14}$ & $\underline{95.42}_{\pm1.41}$ \\

\rowcolor{blue!15}
Ours & $77.34_{\pm0.22}$ & $\textbf{90.57}_{\pm0.21}$ & $\textbf{97.25}_{\pm0.07}$
& $77.30_{\pm0.61}$ & $\textbf{90.99}_{\pm0.27}$ & $\textbf{97.59}_{\pm0.16}$
& $63.40_{\pm0.36}$ & $\textbf{93.07}_{\pm0.47}$ & $\textbf{98.91}_{\pm0.21}$
& $66.80_{\pm1.04}$ & $\textbf{91.46}_{\pm0.37}$ & $\textbf{98.59}_{\pm0.11}$
& $66.00_{\pm0.70}$ & $\textbf{93.04}_{\pm0.29}$ & $\textbf{98.93}_{\pm0.08}$ \\

\midrule

\multicolumn{16}{c}{\textbf{Objects365 benchmark}} \\

NLL & $28.23_{\pm0.43}$ & $62.90_{\pm3.42}$ & $60.80_{\pm2.95}$
& $29.12_{\pm0.22}$ & $59.40_{\pm2.46}$ & $61.00_{\pm3.03}$
& $26.40_{\pm1.01}$ & $52.99_{\pm4.31}$ & $65.80_{\pm2.14}$
& $32.33_{\pm0.49}$ & $59.19_{\pm1.24}$ & $75.11_{\pm0.98}$
& $28.65_{\pm0.26}$ & $57.95_{\pm3.88}$ & $71.99_{\pm2.91}$ \\

Entropy & $28.23_{\pm0.43}$ & $63.30_{\pm4.91}$ & $60.90_{\pm4.22}$
& $29.12_{\pm0.22}$ & $59.10_{\pm2.42}$ & $60.40_{\pm2.90}$
& $26.40_{\pm1.01}$ & $53.22_{\pm4.23}$ & $65.79_{\pm2.17}$
& $32.33_{\pm0.49}$ & $60.27_{\pm1.32}$ & $75.48_{\pm0.97}$
& $28.65_{\pm0.26}$ & $57.06_{\pm3.12}$ & $74.80_{\pm2.51}$ \\

Internal Conf. & $28.23_{\pm0.43}$ & $68.70_{\pm2.67}$ & $67.40_{\pm2.41}$
& $29.12_{\pm0.22}$ & $65.50_{\pm2.44}$ & $70.00_{\pm2.19}$
& $26.40_{\pm1.01}$ & $\underline{71.74}_{\pm3.10}$ & $78.32_{\pm2.35}$
& $32.33_{\pm0.49}$ & $61.35_{\pm3.62}$ & $77.10_{\pm2.93}$
& $28.65_{\pm0.26}$ & $64.10_{\pm3.34}$ & $79.95_{\pm2.09}$ \\

SVAR & $28.23_{\pm0.43}$ & $64.90_{\pm2.88}$ & $66.60_{\pm0.2.39}$
& $29.12_{\pm0.22}$ & $63.50_{\pm3.22}$ & $68.20_{\pm2.86}$
& $26.40_{\pm1.01}$ & $70.33_{\pm1.56}$ & $\underline{79.02}_{\pm0.42}$
& $32.33_{\pm0.49}$ & $\underline{65.11}_{\pm0.63}$ & $\underline{80.75}_{\pm0.99}$
& $28.65_{\pm0.26}$ & $\underline{72.59}_{\pm2.93}$ & $\underline{80.88}_{\pm2.05}$ \\

Contextual Lens & $28.23_{\pm0.43}$ & $63.20_{\pm3.47}$ & $62.60_{\pm3.06}$
& $29.12_{\pm0.22}$ & $62.10_{\pm1.74}$ & $65.60_{\pm1.25}$
& $26.40_{\pm1.01}$ & $65.81_{\pm2.11}$ & $75.82_{\pm2.26}$
& $32.33_{\pm0.49}$ & $62.57_{\pm4.12}$ & $79.72_{\pm2.03}$
& $28.65_{\pm0.26}$ & $55.31_{\pm3.96}$ & $68.70_{\pm2.91}$ \\

GLSIM & $28.23_{\pm0.43}$ & $\underline{72.60}_{\pm2.58}$ & $\underline{74.60}_{\pm2.42}$
& $29.12_{\pm0.22}$ & $\underline{70.40}_{\pm1.55}$ & $\underline{74.00}_{\pm1.87}$
& $26.40_{\pm1.01}$ & $69.76_{\pm2.26}$ & $80.07_{\pm2.78}$
& $32.33_{\pm0.49}$ & $62.36_{\pm0.69}$ & $79.70_{\pm0.80}$
& $28.65_{\pm0.26}$ & $65.30_{\pm2.91}$ & $75.30_{\pm2.76}$ \\

\rowcolor{blue!15}
Ours & $27.90_{\pm0.35}$ & $\textbf{77.89}_{\pm0.23}$ & $\textbf{84.56}_{\pm0.45}$
& $29.20_{\pm0.28}$ & $\textbf{77.69}_{\pm0.83}$ & $\textbf{85.36}_{\pm0.91}$
& $26.70_{\pm0.31}$ & $\textbf{80.15}_{\pm0.56}$ & $\textbf{87.58}_{\pm0.17}$
& $32.30_{\pm0.57}$ & $\textbf{71.54}_{\pm0.51}$ & $\textbf{84.56}_{\pm0.89}$
& $28.70_{\pm0.61}$ & $\textbf{82.14}_{\pm0.12}$ & $\textbf{88.31}_{\pm0.39}$ \\

\bottomrule
\end{tabular}
}
\end{table}

\begin{table}[h]
\centering
\caption{Performance comparison on CLEVR and SpatialMQA benchmarks.}
\label{tab:attribute_results_with_std}
\setlength{\tabcolsep}{3.2pt}

\resizebox{\linewidth}{!}{
\begin{tabular}{l ccc ccc ccc ccc ccc}
\toprule

\multirow{3}{*}{\textbf{Method}}
& \multicolumn{15}{c}{\textbf{Models}} \\

\cmidrule(lr){2-16}
& 
\multicolumn{3}{c}{\textbf{LLaVA-1.5-7B}}
& \multicolumn{3}{c}{\textbf{LLaVA-1.5-13B}}
& \multicolumn{3}{c}{\textbf{LLaVA-NeXT-13B}}
& \multicolumn{3}{c}{\textbf{Qwen2.5-VL-7B}}
& \multicolumn{3}{c}{\textbf{LLaVA-OneVision1.5-8B}} \\

& 
ACC & AUROC & AUPR
& ACC & AUROC & AUPR
& ACC & AUROC & AUPR
& ACC & AUROC & AUPR
& ACC & AUROC & AUPR \\
\midrule

\multicolumn{16}{c}{\textbf{CLEVR benchmark}} \\

NLL & $43.43_{\pm0.95}$ & $45.22_{\pm1.86}$ & $38.13_{\pm0.91}$
& $47.03_{\pm0.82}$ & $44.42_{\pm0.73}$ & $41.30_{\pm1.00}$
& $49.70_{\pm0.75}$ & $52.56_{\pm1.42}$ & $49.44_{\pm2.21}$
& $97.70_{\pm0.18}$ & $\underline{88.72}_{\pm0.81}$ & $\underline{99.64}_{\pm0.02}$
& $97.98_{\pm0.41}$ & $89.39_{\pm1.86}$ & $99.74_{\pm0.03}$ \\

Entropy & $43.43_{\pm0.95}$ & $44.97_{\pm1.97}$ & $38.02_{\pm0.83}$
& $47.03_{\pm0.82}$ & $46.79_{\pm4.44}$ & $42.64_{\pm4.83}$
& $49.70_{\pm0.75}$ & $51.71_{\pm2.35}$ & $49.11_{\pm2.16}$
& $97.70_{\pm0.18}$ & $88.54_{\pm0.66}$ & $99.63_{\pm0.01}$
& $97.98_{\pm0.41}$ & $\underline{90.29}_{\pm1.75}$ & $\underline{99.76}_{\pm0.03}$ \\

Internal Conf. & $43.43_{\pm0.95}$ & $37.78_{\pm1.08}$ & $35.67_{\pm0.71}$
& $47.03_{\pm0.82}$ & $45.18_{\pm3.41}$ & $47.25_{\pm3.95}$
& $49.70_{\pm0.75}$ & $42.79_{\pm4.21}$ & $44.01_{\pm3.87}$
& $97.70_{\pm0.18}$ & $32.77_{\pm5.89}$ & $95.24_{\pm0.84}$
& $97.98_{\pm0.41}$ & $35.98_{\pm5.72}$ & $96.68_{\pm0.66}$ \\

SVAR & $43.43_{\pm0.95}$ & $55.39_{\pm1.14}$ & $49.13_{\pm1.05}$
& $47.03_{\pm0.82}$ & $50.36_{\pm0.81}$ & $50.12_{\pm1.99}$
& $49.70_{\pm0.75}$ & $53.12_{\pm1.49}$ & $52.31_{\pm1.25}$
& $97.70_{\pm0.18}$ & $73.69_{\pm2.42}$ & $97.31_{\pm0.43}$
& $97.98_{\pm0.41}$ & $75.75_{\pm3.28}$ & $99.30_{\pm0.11}$ \\

Contextual Lens & $43.43_{\pm0.95}$ & $57.16_{\pm0.23}$ & $49.28_{\pm0.06}$
& $47.03_{\pm0.82}$ & $\underline{58.03}_{\pm0.51}$ & $52.77_{\pm0.26}$
& $49.70_{\pm0.75}$ & $58.08_{\pm0.37}$ & $57.97_{\pm1.22}$
& $97.70_{\pm0.18}$ & $82.37_{\pm1.31}$ & $99.47_{\pm0.10}$
& $97.98_{\pm0.41}$ & $71.45_{\pm1.64}$ & $98.94_{\pm0.11}$ \\

GLSIM & $43.43_{\pm0.95}$ & $\underline{59.89}_{\pm0.82}$ & $\underline{50.94}_{\pm1.12}$
& $47.03_{\pm0.82}$ & $57.96_{\pm0.89}$ & $\underline{54.45}_{\pm2.12}$
& $49.70_{\pm0.75}$ & $\underline{60.45}_{\pm0.41}$ & $\underline{61.42}_{\pm0.83}$
& $97.70_{\pm0.18}$ & $79.72_{\pm1.75}$ & $99.26_{\pm0.21}$
& $97.98_{\pm0.41}$ & $82.60_{\pm2.38}$ & $99.31_{\pm0.23}$ \\

\rowcolor{blue!15}
Ours & $43.73_{\pm0.94}$ & $\textbf{63.34}_{\pm0.48}$ & $\textbf{53.85}_{\pm0.41}$
& $47.15_{\pm1.42}$ & $\textbf{60.87}_{\pm1.12}$ & $\textbf{54.57}_{\pm0.42}$
& $48.52_{\pm0.76}$ & $\textbf{63.79}_{\pm0.39}$ & $\textbf{69.76}_{\pm0.53}$
& $97.96_{\pm0.43}$ & $\textbf{92.11}_{\pm0.32}$ & $\textbf{99.77}_{\pm0.43}$
& $97.92_{\pm0.11}$ & $\textbf{92.33}_{\pm0.22}$ & $\textbf{99.78}_{\pm0.01}$ \\

\midrule

\multicolumn{16}{c}{\textbf{SpatialMQA benchmark}} \\

NLL & 30.41 & \underline{62.42} & 40.14
& 32.69 & 57.76 & 41.89
& 34.65 & 60.46 & 46.61
& 36.90 & 59.88 & 51.92
& 35.73 & 60.40 & 48.09 \\

Entropy & 30.41 & 62.33 & \underline{41.84}
& 32.69 & 51.73 & 36.12
& 34.65 & 56.68 & 46.07
& 36.90 & 61.72 & \underline{52.86}
& 35.73 & 60.23 & 48.13 \\

Internal Conf. & 30.41 & 46.41 & 28.53
& 32.69 & 51.85 & 36.40
& 34.65 & 56.68 & 46.07
& 36.90 & 51.35 & 37.16
& 35.73 & 51.65 & 38.54 \\

SVAR & 30.41 & 56.40 & 36.07
& 32.69 & \underline{66.61} & \underline{50.32}
& 34.65 & 65.02 & 50.36
& 36.90 & 61.27 & 48.30
& 35.73 & 57.00 & 44.19 \\

Contextual Lens & 30.41 & 62.36 & 41.27
& 32.69 & 62.44 & 45.58
& 34.65 & 64.87 & 52.75
& 36.90 & \underline{64.84} & 52.44
& 35.73 & 61.03 & 48.41 \\

GLSIM & 30.41 & 61.30 & 40.69
& 32.69 & 64.89 & \textbf{51.11}
& 34.65 & \underline{66.49} & \underline{57.55}
& 36.90 & 62.49 & 52.03
& 35.73 & \underline{63.99} & \underline{49.13} \\

\rowcolor{blue!15}
Ours & $31.17_{\pm0.14}$ & $\textbf{71.32}_{\pm0.27}$ & $\textbf{49.71}_{\pm0.62}$
& $32.76_{\pm0.08}$ & $\textbf{67.84}_{\pm0.14}$ & $47.93_{\pm0.33}$
& $34.52_{\pm0.07}$ & $\textbf{67.05}_{\pm0.36}$ & $\textbf{57.94}_{\pm0.44}$
& $36.92_{\pm0.14}$ & $\textbf{76.06}_{\pm0.18}$ & $\textbf{68.80}_{\pm0.23}$
& $36.48_{\pm0.73}$ & $\textbf{76.04}_{\pm1.09}$ & $\textbf{66.60}_{\pm0.69}$ \\

\bottomrule
\end{tabular}
}
\end{table}

\subsection{Limitation}
The hallucination detection performance of our method largely depends on the capability of the fine-tuned LVLM itself. Our approach is based on the assumption that the LM head can preserve the separability between real tokens and hallucinated tokens in the representation space. If the model performs poorly on the target task, this separability between tokens becomes significantly weaker, as the model itself lacks confidence in its generated answers. Consequently, even after fine-tuning, the LM head cannot effectively translate the underlying representations into a reliable correctness signal. As shown in Table~\ref{tab:attribute_results_with_std}, LLaVA-1.5 exhibits poor reasoning performance on CLEVR, and correspondingly achieves low detection performance. In contrast, Qwen2.5-VL and LLaVA-OneVision demonstrate strong reasoning capabilities on CLEVR, and their hallucination detection results are also significantly better.

\subsection{Case study}

\begin{figure}[t]
    \centering
    \includegraphics[width=\linewidth,height=0.8\textheight]{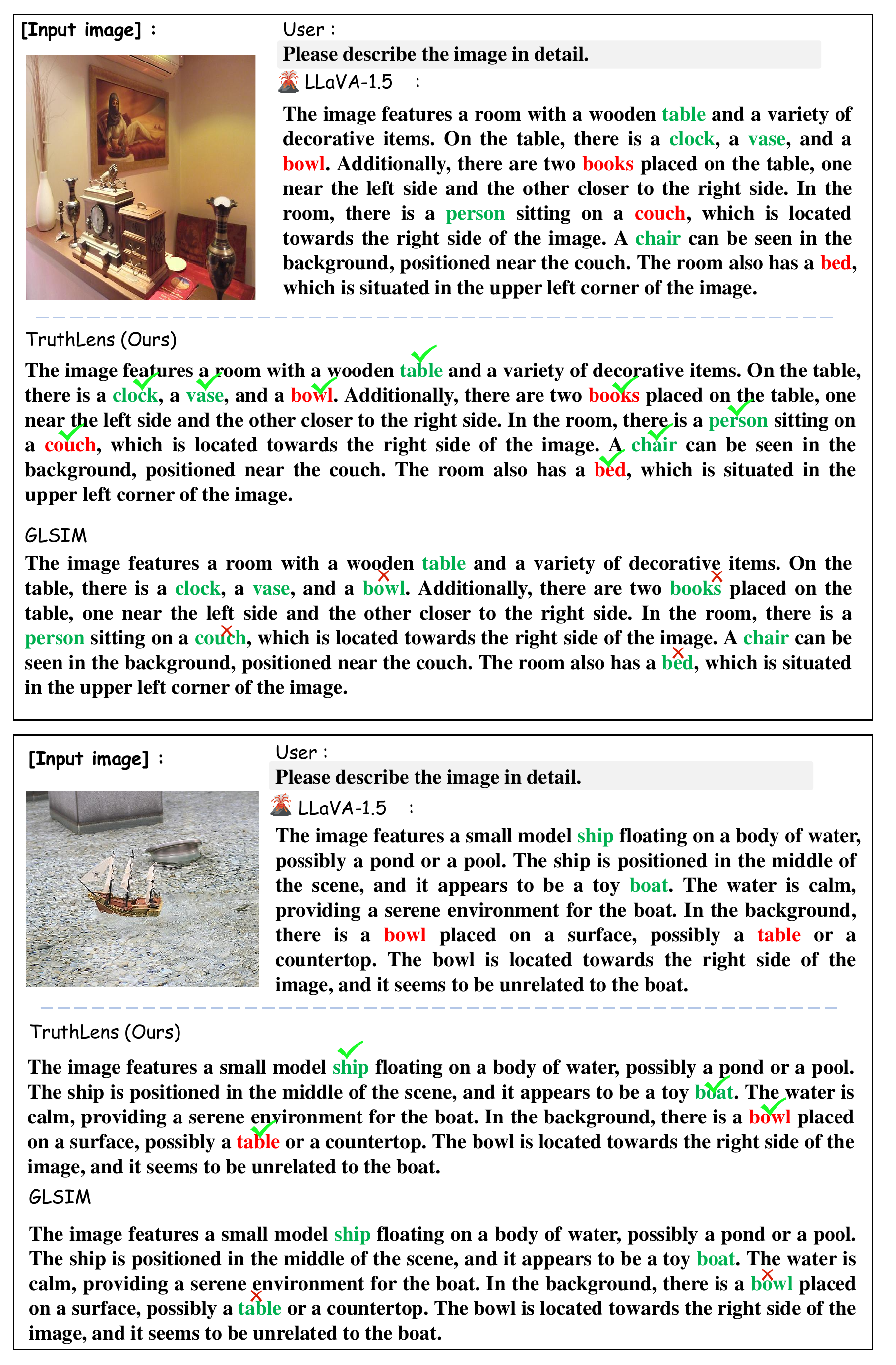}
    \caption{Qualitative comparison of InsLen and GLSIM for detecting object hallucinations, using the LLaVA-1.5-7B model. In the generated responses, ground-truth objects are highlighted in green, while hallucinated objects appear in red. Detection outcomes are color-coded: green indicates correct identification of real objects, orange marks detected hallucinations, and incorrect predictions are flagged with a {\ding{55}}.}
    
\label{fig:case_study}
\end{figure}

We present qualitative examples demonstrating the real-time deployment of TruthLens for object hallucination detection. The decision thresholds were determined on a validation set of 500 samples, set to $0.3$ for TruthLens and $9.3\times10^{-2}$ for GLSIM on the LLaVA-1.5-7B model. As illustrated in Fig.~\ref{fig:case_study}, TruthLens correctly identifies the hallucinated object \textit{couch}, whereas GLSIM fails. Overall, TruthLens exhibits superior reliability in distinguishing hallucinated objects by leveraging the separability between the features of real and hallucinated tokens.

\end{document}